\documentclass[conference]{IEEEtran}

\usepackage{algorithm}
\usepackage{algpseudocode}
\usepackage{mathtools}
\usepackage{multirow}
\usepackage{tabto}

\usepackage{url}
\usepackage{fancyhdr}
\usepackage{hyperref}
\fancypagestyle{firstpage}{
  \fancyhf{}
  
  \fancyfoot[L]{{\em This paper was published in the Proceedings of the 2019 IEEE 13th International 
Conference on Semantic Computing (ICSC), pages 47-54, Newport Beach, CA, USA, 2019. \copyright 2019 IEEE\\
Link to article abstract in IEEE Xplore: http://dx.doi.org/10.1109/ICOSC.2019.8665646}}
}
\begin{document}

\title{Stopping Active Learning based on Predicted Change of F Measure for Text Classification}

\author{\IEEEauthorblockN{Michael Altschuler}
\IEEEauthorblockA{Department of Computer Science\\
The College of New Jersey\\
Ewing, New Jersey 08618\\
Email: altschm1@tcnj.edu}
\and
\IEEEauthorblockN{Michael Bloodgood}
\IEEEauthorblockA{Department of Computer Science\\
The College of New Jersey\\
Ewing, New Jersey 08618\\
Email: mbloodgood@tcnj.edu}
}

\pagenumbering{gobble}

\maketitle

\thispagestyle{firstpage}

\begin{abstract}
During active learning, an effective stopping method allows users to limit the number of annotations, which is cost effective.  In this paper, a new stopping method called Predicted Change of F Measure will be introduced that attempts to provide the users an estimate of how much performance of the model is changing at each iteration.  This stopping method can be applied with any base learner. This method is useful for reducing the data annotation bottleneck encountered when building text classification systems.
\end{abstract}

\section{Introduction}

The use of active learning to train machine learning models has been used as a way to reduce annotation costs for text and speech processing applications \cite{hantke2017, bloodgood2010ACL, lee2012, mairesse2010, miura2016}.  Active learning has been shown to have a particularly large potential for reducing annotation cost for text classification \cite{lewis1994, tong2001}. Text classification is one of the most important fields in semantic computing and it has been used in many applications \cite{mishler2017ICSC, hoi2006, janik2008, allahyari2014, kanakaraj2015}.

\subsection{Active Learning}

Active learning is a form of machine learning that gives the model the ability to select the data on which it wants to learn from and to choose when to end the process of training.

In active learning, the model is first provided a small batch of annotated data to be trained on.  Then, in each following iteration, the model selects a small batch and removes this batch from a large unlabeled set of examples.  This batch is then annotated and added to the labeled set of examples. Models are trained at each iteration on the growing set of labeled data. By carefully selecting the batches of data to label at each iteration, learning efficiency can be increased. However, in order to enable the potential benefits enabled by active learning, a method for stopping the process must be developed.

Active learning is most applicable in fields where there is a lot of unlabeled data readily available, but the cost of annotations is very high. Ideally, the model is able to maximize its performance on the least amount of data to minimize the annotation cost with active learning.  The model needs to correctly identify the batch that will increase the performance of the model by the most and correctly identify the optimal time to end the active learning process.

In previous papers, there has been extensive research on what is the optimal batch to select.  Lewis and Gale proposed a method called uncertainty sampling where examples are chosen where the model is least certain of classification \cite{lewis1994}.  Schohn and Cohn have developed an uncertainty sampling method for support vector machines, which can be called closest-to-the-hyperplane, where examples for the batch are chosen based on the distance from the hyperplane \cite{schohn2000}.  Tang, Luo, and Roukos developed an uncertainty sampling method based on the entropy formula \cite{tang2002}.  Another technique, called query by committee, involves selecting the examples based on how much a set of models disagree on the output of an example \cite{seung1992}.  However, closest-to-hyperplane selection has been found to be superior to query-by-committee \cite{bloodgood2018ICSC}; hence, we use closest-to-hyperplane selection in our experiments.  Konyushkova, Raphael, and Fua propose a sampling method that selects the batch that reduces the expected error of a model \cite{konyushkova2017}. The focus of this paper will be on the stopping method.

\subsection{Stopping Method}

A stopping method essentially tells the model when to end the process of active learning. Using the terminology introduced in \cite{bloodgood2009}, a stopping method is considered {\em aggressive} if a stopping method prioritizes minimizing annotation costs over achieving maximum performance.  Hence an aggressive stopping method will stop relatively earlier.  A stopping method is considered {\em conservative} if a stopping method prioritizes maximizing performance over minimizing annotation costs.  Hence a conservative method will stop relatively later.

The benefits of an effective stopping method are two-fold.  First, an effective stopping method can limit annotation cost ideally without compromising performance.  This can be economically beneficial in many applications.  Second, there are reported instances where a model trained on some data is actually superior to a model trained on all the data \cite{schohn2000}.

Evaluating a stopping criterion can often be a subjective issue that is determined by many factors.  This is because users may prefer a conservative or aggressive stopping method based on the application.  In a cost-sensitive project where performance maximization isn't critical, an aggressive stopping method may be preferred.  In a domain where performance is the priority, a conservative stopping method may be preferred.

\section{Related Work}

Various stopping methods have been proposed over the years. In 2000, Schohn and Cohn recommended to end the active learning process once all the examples within the margins of an SVM model have been labeled \cite{schohn2000}.  The reasoning behind this stopping method was that the examples most likely to change the hyperplane of a support vector machine lie within the margins.  Once the margin has been exhausted, all the useful examples have been labeled and active learning can be stopped.

In 2007 and 2008, Zhu, Wang, and Hovy proposed several stopping methods.  In 2007, Zhu and Hovy proposed two stopping methods (max-confidence and min-error) \cite{zhu2007}.  Max-confidence stops when the entropy on each selected unlabeled example is less than a threshold.  Min-error stops when the model's predictions on the unlabeled batch exceeds a certain performance.  In 2008, Zhu, Wang, and Hovy proposed a new strategy called the minimum expected error strategy that stops when the total expected error on the unlabeled pool is less than a certain threshold \cite{zhu2008learning}.  Zhu, Wang, and Hovy also proposed a combination of the previous strategies with the classification change stopping method which stops when no change of classification of an unlabeled example happens on consecutive iterations \cite{zhu2008}.

In 2008, Laws and Schutze proposed two stopping methods called performance convergence and uncertainty convergence \cite{laws2008}.  Performance convergence stops when the estimation of accuracy of the unlabeled pool converges so that the gradient of performance estimates is below a certain threshold. Uncertainty convergence stops when the uncertainty of the last selected instance of the batch has an uncertainty measure below a minimum threshold. For support vector machines and other learners that do not provide probabilities of classification, only uncertainty convergence can be used.

In 2008, Vlachos proposed a stopping criterion that stops when the confidence on a set of examples consistently drops \cite{vlachos2008}.  The logic in this stopping method is that as you add more labeled data to the model, the confidence of the model will initially rise, but then decline after the performance stabilized.  However, in experiments, it has been found that this assumption does not always hold.

In 2009, Bloodgood and Vijay-Shanker proposed a stopping method called stabilizing predictions \cite{bloodgood2009}.  This stopping method takes a set of data that does not have to be labeled called a stop set and classifies each example in the stop set as positive or negative at each iteration of active learning.  If the classifications of the examples in the stop set begin to stabilize so that the agreement between the classifications of successively learned models across several iterations is above a threshold as measured by Cohen's Kappa \cite{Cohen1960}, then the active learning process ends.  This stopping method is easy-to-implement, applicable with any base learner, and the number of iterations and the Kappa threshold can be altered depending on how conservative or aggressive one chooses to be. 

In 2010, Ghayoomi proposed a stopping criterion called the extended variance model \cite{ghayoomi2010}.  The extended variance model stops once the variance of the confidence of the unlabeled pool decreases by a minimum threshold over a certain number of iterations.  However, in our experiments with text classification (see Section~\ref{h:AccuracyPredictingChangeF}), this stopping method does not stop.

In 2013, Bloodgood and Grothendieck were able to prove that when Cohen's Kappa is greater than or equal to some threshold T, then the change of F measure is less than or equal to $\frac{4(1-T)}{T}$ \cite{bloodgood2013}.  This means that stabilizing predictions provides an intuition of how much performance will change on the stop set. Most recently, Beatty and others \cite{beatty2019} were able to show that stopping methods based on unlabeled data such as the stabilizing predictions method of \cite{bloodgood2009, bloodgood2013} are more effective than stopping methods based on labeled data. 

\section{Predicted Change of F} \label{stopping_method}

As stated earlier, determining how effective a stopping method is can be subjective as one does not know how aggressive or conservative the user wants to be.  Therefore, an ideal stopping method should accomplish the following goals:
\begin{enumerate}
  \item The user should have the ability to determine how aggressive or conservative the stopping method should be based on a given set of parameters.
  \item The stopping method should provide a sense of intuition on how much performance was gained or lost on the previous iteration so that the user can effectively determine if the marginal benefit of the change in performance will outweigh the cost of annotating another batch.
  \item The stopping method should be able to work with various learners.
\end{enumerate}

Hypothetically, one could calculate the change in performance by creating a separate test set of examples with the labels provided called the validation set and calculate the change in performance from this set.  However, this idea brings about two problems.  First, if the validation set is really small, there are no guarantees that this validation set will be an accurate representation of the population.  Second, if the validation set is really large, the annotation costs for providing the labels will be really high which defeats the whole purpose of active learning.  Therefore, another method is preferred that can determine when to stop without requiring any additional labeled data.

In 2009, Bloodgood and Vijay-Shanker proposed a stopping method that stops when the predictions at different iterations stabilize \cite{bloodgood2009}.  This method uses a set of data that does not need to be labeled called the stop set.  At each iteration, the model classifies each example in the stop set as positive or negative.  After three consecutive iterations, if the measurement of agreement between the three iterations as measured by the Kappa value is above a threshold, active learning ends.  In other words, once the predictions on the stop set stabilize, no more examples are annotated.  The advantage of this approach is that it is applicable to any base learner, offers the user the flexibility to adjust the threshold, and was shown to have stable strong performance over many datasets.

In 2013, Bloodgood and Grothendieck proved that when the Kappa value is greater than or equal to any threshold T, then the change of F measure is less than or equal to $\frac{4(1-T)}{T}$ \cite{bloodgood2013}.  This theorem provides an intuition of how much performance will change on the stop set, which is valuable information to the user.  F measure was used over accuracy because F measure is commonly used in text classification and other information extraction tasks. F measure is defined as the harmonic mean of precision and recall.  Precision is the percentage of system-proposed positives that are truly positive. Recall is the percentage of truly positive examples that the system proposed as positive.

The result from \cite{bloodgood2013} only provides an upper bound. In practice, the difference in performance could be much smaller. It would be an advantage to the user to provide a more accurate estimate of the change of F measure.  To do this, a more direct conversion to change of performance is needed.

In 2013, Bloodgood and Grothendieck described a contingency table that compares the predictions made by $M_t$ and $M_{t-1}$ where $M_t$ is the model that was learned at the most recent iteration and $M_{t-1}$ is the model that was learned at the previous iteration\cite{bloodgood2013}.  As seen in Table~\ref{t:contingency_table_1}, when $M_t$ and $M_{t-1}$ both label the example as positive, that example is added to the count of \textit{a}.  When $M_t$ and $M_{t-1}$ both label the example as negative, that example is added to the count of \textit{d}.  When $M_t$ labels the example as positive and $M_{t-1}$ labels the example as negative, that example is added to the count of \textit{c}.  When $M_t$ labels the example as negative and $M_{t-1}$ labels the example as positive, that example is added to the count of \textit{b}.

\begin{table}[t]
\caption{A contingency table of labeled examples at the most recent iteration and the previous iteration.}
\label{t:contingency_table_1}
\centering
\begin{tabular}{  c  c | c | c | c }
 \ & \ &  \multicolumn{2}{c}{$M_t$}  & \ \\
  \ & \ & + & - & Total \\
  \hline
  \multirow{2}{*}{$M_{t - 1}$} & + & a & b & a + b \\
   \ & - & c & d & c + d \\
  \hline
  \ & Total & a + c & b + d & n \\
\end{tabular}
\end{table}

However, we only know the system-proposed labels of examples in the stop set.  Therefore, Bloodgood and Grothendieck described the following two tables (Table~\ref{t:contingency_table_2} and Table~\ref{t:contingency_table_3}) which show the truly positive and negative examples of \textit{a}, \textit{b}, \textit{c}, and \textit{d}.  Let $a_+$ be the number of examples contributing to the count of \textit{a} that are truly positives and let $a_-$ be the examples contributing to the count of \textit{a} that are truly negative.  Let the same apply for \textit{b}, \textit{c}, and \textit{d}.

\begin{table}[t]
\caption{A contingency table of labeled examples that are truly positive at the most recent iteration and the previous iteration.}
\label{t:contingency_table_2}
\centering
\begin{tabular}{  c  c | c | c | c }
 \ & \ &  \multicolumn{2}{c}{$M_t$}  & \ \\
  \ & \ & + & - & Total \\
  \hline
  \multirow{2}{*}{$M_{t - 1}$} & + & $a_{+}$ & $b_{+}$ & $a_{+} + b_{+}$ \\
   \ & - & $c_+$ & $d_+$ & $c_+ + d_+$ \\
  \hline
  \ & Total & $a_+ + c_+$ & $b_+ + d_+$ & $n_+$ \\
\end{tabular}
\end{table}

\begin{table}[t]
\caption{A contingency table of labeled examples that are truly negative at the most recent iteration and the previous iteration.}
\label{t:contingency_table_3}
\centering
\begin{tabular}{  c  c | c | c | c }
 \ & \ &  \multicolumn{2}{c}{$M_t$}  & \ \\
  \ & \ & + & - & Total \\
  \hline
  \multirow{2}{*}{$M_{t - 1}$} & + & $a_{-}$ & $b_{-}$ & $a_{-} + b_{-}$ \\
   \ & - & $c_-$ & $d_-$ & $c_- + d_-$ \\
  \hline
  \ & Total & $a_- + c_-$ & $b_- + d_-$ & $n_-$ \\
\end{tabular}
\end{table}

The contingency tables of $M_t$ versus ground truth and $M_{t-1}$ versus ground truth are required if the true change of F performance needs to be calculated.  Table~\ref{t:contingency_table_4} shows $M_t$ versus ground truth and Table~\ref{t:contingency_table_5} shows $M_{t-1}$ versus ground truth.

\begin{table}[t]
\caption{A contingency table of labeled examples at the most recent iteration versus the ground truth.}
\label{t:contingency_table_4}
\centering
\begin{tabular}{  c  c | c | c | c }
 \ & \ &  \multicolumn{2}{c}{$M_t$}  & \ \\
  \ & \ & + & - & Total \\
  \hline
  \multirow{2}{*}{$Truth$} & + & $a_{+} + c_+$ & $b_{+} + d_+$ & $n_+$ \\
   \ & - & $a_- + c_-$ & $b_{-} + d_-$ & $n_-$ \\
  \hline
  \ & Total & $a + c$ & $b + d$ & $n$ \\
\end{tabular}
\end{table}

\begin{table}[t]
\caption{A contingency table of labeled examples at the previous iteration versus the ground truth.}
\label{t:contingency_table_5}
\centering
\begin{tabular}{  c  c | c | c | c }
 \ & \ &  \multicolumn{2}{c}{$M_{t-1}$}  & \ \\
  \ & \ & + & - & Total \\
  \hline
  \multirow{2}{*}{$Truth$} & + & $a_{+} + b_+$ & $c_{+} + d_+$ & $n_+$ \\
   \ & - & $a_- + b_-$ & $c_{-} + d_-$ & $n_-$ \\
  \hline
  \ & Total & $a + b$ & $c + d$ & $n$ \\
\end{tabular}
\end{table}

To calculate the F measure of the model from the contingency counts, the formula is:

\begin{equation} F = \frac{2a}{2a + b + c} \label{e:f_measure} \end{equation}

The formula for the change in F measure in terms of $a_+$, $a_-$, $b_+$, $b_-$, $c_+$, $c_-$, $d_+$, and $d_-$ is:

\begin{equation} \begin{aligned} \Delta F = \frac{2(a_+ + c_+)}{2(a_+ + c_+) + b_+ + d_+ + a_- + c_-}\\ - \frac{2(a_+ + b_+)}{2(a_+ + b_+) + c_+ + d_+ + a_- + b_-} \end{aligned} \label{e:true_f_measure} \end{equation}

However, one does not know the true values of $a_+$, $a_-$, $b_+$, $b_-$, $c_+$, $c_-$, $d_+$, and $d_-$.  Therefore, a method is needed to estimate the values of $a_+$, $a_-$, $b_+$, $b_-$, $c_+$, $c_-$, $d_+$, and $d_-$.  For estimating the values, the following assumption will be made: the model at each iteration will get progressively better as it is trained on more labeled data.  Therefore, on all cases where $M_{t-1}$ and $M_{t}$ disagree, we will assume that $M_t$ is correct.  In other words, we assume $a_+ = a$, $a_- = 0$, $b_+ = 0$, $b_- = b$, $c_+ = c$, $c_- = 0$, $d_+ = 0$, and $d_- = d$.  In Section~\ref{h:Assumption}, we evaluate the effectiveness of this assumption.

The predicted change of F can now be described as:

\begin{equation} \Delta \hat{F} = \frac{2(a + c)}{2(a + c)} - \frac{2a}{2a + c + b} = 1 - \frac{2a}{2a + b + c} \label{e:pred_change_f} \end{equation}

When $\Delta \hat{F} < \epsilon$ where $\epsilon$ is a predefined threshold over $k$ windows, the active learning process stops.  Or in other words:

\[
  StoppingMethod = 
    \begin{cases}
      Stop,  & \text{if $\Delta \hat{F} < \epsilon$} \\ &   \text{over $k$ windows}
    \\[4pt]
      Continue, & \text{otherwise.}
    \end{cases}
\]

This stopping method fulfills all three goals of an effective stopping method.  First, the user has the ability to determine how aggressive or conservative the stopping method should be based on the parameters $k$ and $\epsilon$.  Second, the stopping method provides an estimate of how much the performance of the model will change at each iteration by $\Delta \hat{F}$.  These experiments will be shown in Section~\ref{h:Accuracy}.  Third, the stopping method is able to work with any learner as only the predictions on a stop set of data are needed.

\section{Experiments}

Four experiments will be run regarding predicted change of F:

\begin{enumerate}
  \item First, we will do an analysis of the assumption that $M_t$ is more accurate than $M_{t-1}$ by looking at what proportion of $a$ is truly positive (ideally approximately 1), what proportion of $b$ is truly positive (ideally approximately 0), what proportion of $c$ is truly positive (ideally approximately 1), and what proportion of $d$ is truly positive (ideally approximately 0).
  \item Second, the predicted change of F measure will be compared to the change of performance on a separate labeled test set.  In this way, the accuracy of this new measurement can be evaluated and will also be compared to the threshold defined by the Kappa variable.  This will also be compared to the upper bound proven by Bloodgood and Grothendieck \cite{bloodgood2013}.
  \item Third, predicted change of F will be compared to previous stopping methods to evaluate how aggressive/conservative the stopping method is.
  \item Fourth, an analysis on the parameters will be done to investigate how the change in $\epsilon$ and $k$ will change stopping behavior.
\end{enumerate}

\subsection{Experimental Setup}
Predicted change of F was evaluated in the domain of binary text classification.  The datasets used were 20Newsgroups, Reuters, and WebKB.  20Newsgroups contains 20 different categories of articles such as atheism, space, or baseball.  The average for 20Newsgroups was done by averaging the 20 categories.  For Reuters, 10 categories were used that included topics such as corn, money, and wheat.  This is consistent with past practices \cite{joachims1998, dumais1998}. The average was calculated by averaging the ten categories.  The four categories, students, faculty, project, and course, were used for WebKB.  This is also consistent with past practices \cite{mccallum1998, zhu2008, zhu2008learning}.  Ten folds were used, and the average was calculated for each category.  All datasets were downloaded July 13, 2017.

The base learner used for these experiments was a support vector machine.  A support vector machine defines a hyperplane that divides examples into two categories (+1 or -1).  The decision function is defined in terms of $f(x)=sgn(w*x + b)$.
Support vector machines were used because they have good performance in the field of text classification and a lot of past analysis of stopping methods has been done with support vector machines.

The dataset from 20Newsgroup includes 11314 articles of 20 different categories. The dataset from Reuters contains 7780 articles of 10 different categories.  The dataset from WebKB contains 7445 articles of four different categories. We used a batch size of 0.5\% of the size of the initial unlabeled pool and a stop set size of 50\% of the size of the initial unlabeled pool for all the datasets. A larger batch size can increase computational tractability, however, it also degrades the effectiveness of stopping methods \cite{beatty2018ICSC}. This is why we used a batch size of 0.5\% of the initial unlabeled pool in all our experiments. 

In our experiments, the selection algorithm selects the batch of examples that are closest to the hyperplane \cite{tong2002, tong2001, campbell2000, schohn2000}.  Recently this selection algorithm was found to be superior to query by committee selection \cite{bloodgood2018ICSC}. All parameters used for other stopping methods were as recommended in previous papers \cite{schohn2000, zhu2008, laws2008, vlachos2008, bloodgood2009, ghayoomi2010, bloodgood2013}.

\subsection{Analysis of Assumption} \label{h:Assumption}

In Section~\ref{stopping_method}, the assumption was made that in areas of disagreement, the more recent iteration will generally be superior because it is trained on more data.  Therefore, $a_+ = a$, $a_- = 0$, $b_+ = 0$, $b_- = b$, $c_+ = c$, $c_- = 0$, $d_+ = 0$, and $d_- = d$.  However, it is fair to assume that not all of the examples labeled by consecutive iterations as positive ($a$) will have a ground-truth value of positive.  Likewise, this holds true for $b$, $c$, and $d$.

Therefore, the goal of this experiment is to find out what proportion of examples labeled a are truly positive, what proportion of examples labeled b are truly positive, what proportion of examples labeled c are truly positive, and what proportion of examples labeled d are truly positive.

Let $P(+ | a)$ be the conditional probability that an example has a ground truth value of positive given an example is labeled by $M_t$ and $M_{t-1}$ as positive.
Let $P(+ | b)$ be the conditional probability that an example has a ground truth value of positive given an example is labeled by $M_t$ as negative and by $M_{t-1}$ as positive.
Let $P(+ | c)$ be the conditional probability that an example has a ground truth value of positive given an example is labeled by $M_t$ as positive and by $M_{t-1}$ as negative.
Let $P(+ | d)$ be the conditional probability that an example has a ground truth value of positive given an example is labeled by $M_t$ and $M_{t-1}$ as negative.

If the assumptions were perfect, $P(+ | a)$ would be 1; $P(+ | b)$ would be 0; $P(+ | c)$ would be 1; $P(+ | d)$ would be 0.

The values of $P(+ | a)$, $P(+ | b)$, $P(+ | c)$, and $P(+ | d)$ at different stopping points for each dataset are shown in Table~\ref {t:assumption}.

\begin{table}[t]
\caption{This table shows the proportion of examples labeled $a$ that are truly positive, the proportion of examples labeled $b$ that are truly positive, the proportion of examples labeled $c$ that are truly positive, and the proportion of examples labeled $d$ that are truly positive for all datasets.  Each conditional probability is calculated at different stopping points for when $\epsilon = 0.01$, $\epsilon = 0.05$ and $\epsilon = 0.1$.}
\label{t:assumption}
\centering
\begin{tabular}{| c | c | c | c | c | c | }
\hline
﻿Dataset & $\epsilon$ & $P(+ | a)$ & $P(+ | b)$ & $P(+ | c)$ & $P(+ | d)$ \\
\hline
20Newsgroups & 0.01 & 99.4\% & 3.70\% & 97.3\% & 0.497\% \\
\hline
Reuters & 0.01 & 89.0\% & 20.9\% & 64.9\% & 0.354\% \\
\hline
WebKB - faculty & 0.01 & 96.4\% & 12.7\% & 88.8\% & 1.17\% \\
\hline
WebKB - project & 0.01 & 98.7\% & 0.00\% & 100\% & 1.38\% \\
\hline
WebKB - student & 0.01 & 93.5\% & 17.5\% & 83.0\% & 2.25\% \\
\hline
WebKB - course & 0.01 & 95.3\% & 13.0\% & 93.0\% & 1.07\% \\
\hline
\hline
20Newsgroups & 0.05 & 98.0\% & 15.0\$ & 74.9\% & 0.808\% \\
\hline
Reuters & 0.05 & 86.9\% & 29.4\% & 48.0\% & 1.01\% \\
\hline
WebKB - faculty & 0.05 & 91.9\% & 27.4\% & 58.2\% & 1.84\% \\
\hline
WebKB - project & 0.05 & 95.1\% & 9.45\% & 73.8\% & 2.11\% \\
\hline
WebKB - student & 0.05 & 87.4\% & 30.8\% & 52.6\% & 3.59\% \\
\hline
WebKB - course & 0.05 & 91.3\% & 32.0\% & 60.7\% & 1.73\% \\
\hline
\hline
20Newsgroups & 0.10 & 96.8\% & 17.3\% & 69.6\% & 1.15\% \\
\hline
Reuters & 0.10 & 86.2\% & 19.1\% & 61.7\% & 1.23\% \\
\hline
WebKB - faculty & 0.10 & 89.0\% & 27.4\% & 58.1\% & 2.42\% \\
\hline
WebKB - project & 0.10 & 91.4\% & 22.7\% & 62.5\% & 2.61\% \\
\hline
WebKB - student & 0.10 & 86.0\% & 39.1\% & 55.1\% & 5.01\% \\
\hline
WebKB - course & 0.10 & 89.1\% & 34.7\% & 60.1\% & 2.37\% \\
\hline
\end{tabular}
\end{table}

Several observations can be drawn from this experiment.  First, when $\epsilon$ is 0.01, the assumptions are very accurate.  However, as $\epsilon$ increases, the assumptions become slightly less accurate with regards to $P(+ | b)$ and $P(+ | c)$.  However, these results are overall consistent with the assumption that $a_+ = a$, $a_- = 0$, $b_+ = 0$, $b_- = b$, $c_+ = c$, $c_- = 0$, $d_+ = 0$, and $d_- = d$. 

\subsection{Accuracy of Predicting Change of F Measure} \label{h:AccuracyPredictingChangeF} \label{h:Accuracy}

The goal of the experiment reported in this section is to see how well predicted change of F predicts the true change of F on the test set.  The metric used to evaluate this is the absolute value of the difference between the predicted change of F and the change of F measure on the test set .  More formally, this can be referred to as:
$ D = |predicted\:change\:of\:F - change\:of\:F\:on\:test\:set|$.
Predicted change of F was compared to the upper bound of change of F proven in \cite{bloodgood2013}.  The two tables are shown as Table~\ref{t:PCF} and Table~\ref{t:kappa}.

\begin{table}[t]
\caption{This table shows the absolute value of the difference between the predicted change of F and the actual change of F on the test set.  For WebKB, the average of the 10 folds was used.}
\label{t:PCF}
\centering
\tiny
\begin{tabular}{ | c | c | c | c | c | c | c | c | c | }
\hline
 & \multicolumn{8}{c|}{Iterations}\\
\hline
Dataset & 2 & 3 & 4 & 5 & 6 & 7 & 8 & 9 \\
\hline
20Newsgroups & 0.649 & 0.219 & 0.134 & 0.112 & 0.089 & 0.072 & 0.051 & 0.041 \\
\hline
Reuters & 0.249 & 0.228 & 0.182 & 0.117 & 0.095 & 0.061 & 0.047 & 0.022 \\
\hline
WebKB - course & 0.373 & 0.176 & 0.166 & 0.098 & 0.084 & 0.067 & 0.055 & 0.049 \\
\hline
WebKB - faculty & 0.325 & 0.224 & 0.155 & 0.109 & 0.069 & 0.086 & 0.054 & 0.039 \\
\hline
WebKB - project & 0.709 & 0.489 & 0.342 & 0.282 & 0.184 & 0.133 & 0.106 & 0.087 \\
\hline
WebKB - student & 0.407 & 0.220 & 0.148 & 0.140 & 0.111 & 0.088 & 0.074 & 0.090 \\
\hline
\end{tabular}
\end{table}

\begin{table}[t]
\caption{This table shows the absolute difference between the upper bound for change of F on the stop set as proven in \cite{bloodgood2013} and the true change of F on the validation set.  For WebKB, the average of the 10 folds was used.}
\label{t:kappa}
\tiny
\centering
\begin{tabular}{ | c | c | c | c | c | c | c | c | c | }
\hline
\ & \multicolumn{8}{c|}{Iterations}\\
\hline
Dataset & 2 & 3 & 4 & 5 & 6 & 7 & 8 & 9 \\
\hline
20Newsgroups & 1.00 & 1.00 & 1.00 & 0.851 & 0.578 & 0.424 & 0.299 & 0.193 \\
\hline
Reuters & 1.00 & 1.00 & 1.00 & 1.00 & 1.00 & 0.513 & 0.272 & 0.139 \\
\hline
WebKB - course & 1.00 & 1.00 & 1.00 & 0.655 & 0.527 & 0.392 & 0.297 & 0.243 \\
\hline
WebKB - faculty & 1.00 & 1.00 & 1.00 & 0.806 & 0.464 & 0.420 & 0.315 & 0.243 \\
\hline
WebKB - project & 1.00 & 1.00 & 1.00 & 1.00 & 1.00 & 0.932 & 0.719 & 0.586 \\
\hline
WebKB - student & 1.00 & 1.00 & 1.00 & 1.00 & 0.748 & 0.586 & 0.493 & 0.510 \\
\hline
\end{tabular}
\end{table}

There are a few observations that can be made from looking at the results.  First, when there were few annotations labeled, the predicted change of F was inaccurate. However, as the number of annotations increased, the accuracy of the predicted change of F became more accurate. As the predicted change of F converges to 0, the more accurate the method predicts the true change of F. Second, the predicted change of F always provided a more accurate prediction of change of F than if one were to use the upper bound of the change of F as an estimate of the change of F at every iteration on all datasets.

\subsection{Analysis of Aggressiveness of Stopping Method} \label{h:SMComparisons}

The goal of the experiment reported in this section is to see how our new stopping method compares with previous stopping methods.  

We define the optimal ordered pair ($optimalA$, $optimalP$) where $optimalP$ is the maximum performance over the entire learning process minus $\epsilon_2$.  The purpose of $\epsilon_2$ was so that a small increase at the end of the training wouldn't count that iteration as an ideal stopping point.  For all our experiments, $\epsilon_2 = 0.01$.

$optimalA$ is the number of annotations at the earliest iteration that achieves a minimum performance of $optimalP$.  The metrics used to evaluate the aggressiveness of a stopping method are:\\

Percentage of performance achieved:\\
$
P = \frac{performance\:of\:model\:at\:stopping\:point}{optimalP}
$\\

and\\ 

Percentage of annotations used:\\
$
A = \frac{number\:of\:annotations\:used\:at\:stopping\:point}{optimalA}
$\\

The predicted change of F with $\epsilon$ =  0.005 and $k = 1$ (PCF[0.005]) was compared to margin exhaustion (ME) \cite{schohn2000}, drop in confidence (DC) \cite{vlachos2008}, confidence convergence (CC) \cite{laws2008}, min-error with classification change (MECC) \cite{zhu2008}, stabilizing predictions (SP) \cite{bloodgood2009}, and the extended variance model (EVM) \cite{ghayoomi2010}.  Note that all parameters for all methods were kept the same as conducted in the experiments of Bloodgood and Vijay-Shanker \cite{bloodgood2009}.  Table ~\ref{t:performance} and Table ~\ref{t:annotations} show these results.

\begin{table}[t]
\caption{This table shows the average percentage of $optimalP$ achieved among the different stopping methods.  For 20 Newsgroups and Reuters, the average among all the categories were used.  For WebKB, the average of the 10 folds was used.}
\label{t:performance}
\tiny
\centering
\begin{tabular}{ | c | c | c | c | c | c | c | c | }
\hline
﻿ & ME & DC & CC & MECC & SP & EVM & PCF (0.005)\\
\hline
20Newsgroups & 0.982 & 0.586 & 0.879 & 0.973 & 0.976 & 0.987 & 0.973\\
\hline
Reuters & 0.969 & 0.974 & 0.843 & 0.971 & 0.955 & 0.967 & 0.958\\
\hline
WebKB - course & 0.979 & 0.872 & 0.886 & 0.982 & 0.968 & 0.943 & 0.975\\
\hline
WebKB - faculty & 0.972 & 0.895 & 0.923 & 0.973 & 0.975 & 0.954 & 0.975\\
\hline
WebKB - project & 0.957 & 0.556 & 0.672 & 0.725 & 0.891 & 0.911 & 0.906\\
\hline
WebKB - student & 0.980 & 0.877 & 0.831 & 0.991 & 0.978 & 0.971 & 0.976\\
\hline
\end{tabular}
\end{table}

\begin{table}[t]
\caption{This table shows the average percentage of $optimalA$ achieved among the different stopping methods.  For 20Newsgroups and Reuters, the average among all the categories were used.  For WebKB, the average of the 10 folds was used.}
\label{t:annotations}
\tiny
\centering
\begin{tabular}{ | c | c | c | c | c | c | c | c | }
\hline
& ME & DC & CC & MECC & SP & EVM & PCF - 0.005\\
\hline
20Newsgroups & 2.05 & 0.32 & 0.81 & 0.90 & 0.98 & 13.36 & 0.91\\
\hline
Reuters & 1.70 & 6.17 & 0.54 & 1.02 & 1.05 & 14.96 & 1.00\\
\hline
WebKB - course & 2.51 & 11.94 & 0.46 & 1.65 & 1.46 & 15.20 & 1.36\\
\hline
WebKB - faculty & 2.18 & 3.81 & 0.47 & 1.67 & 1.32 & 12.82 & 1.24\\
\hline
WebKB - project & 1.07 & 0.79 & 0.29 & 0.54 & 0.72 & 5.77 & 0.67\\
\hline
WebKB - student & 1.91 & 3.24 & 0.22 & 1.36 & 1.06 & 7.17 & 1.05\\
\hline
\end{tabular}
\end{table}

When looking at the percentage of performance achieved, predicted change of F typically scored very high and achieved at least 90\% of optimal performance on most datasets. When looking at each dataset individually, predicted change of F achieved at least 95\% of optimal performance on all datasets except WebKB - projects.  It is important to note that most of the other stopping methods also struggled to achieve optimal performance on this dataset and category.

When looking at the percentage of annotations used, predicted change of F typically achieved a score very close to 1.00.  This means that predicted change of F was relatively aggressive.  It is also important to note that the parameters $\epsilon = 0.005$ and $k = 1$ can be adjusted to be more or less aggressive depending on the application. In the next subsection we investigate the  impacts that the parameters have on stopping behavior. 

\subsection{Analysis of Parameters}

One of the goals of this section is to analyze and see how a change in $\epsilon$ and the number of windows $k$ will impact the aggressiveness of the stopping method.  The intuition is that a decrease in $\epsilon$ and an increase in $k$ would lead to a more conservative stopping method.  However, to what degree would it be more conservative?  To do this, the number of annotations at each stopping point was found for values of $\epsilon$ in the set  \{0.01, 0.02, 0.03, 0.04, 0.05, 0.06, 0.07, 0.08, and 0.09\} and for different values of $k$ in the set  \{1, 2, 3\}.  Then a regression was done with the model of: $\hat{y} = \beta_1\epsilon + \beta_2k + \beta_3$ where $\hat{y}$ is the predicted number of annotations, $\epsilon$ is the epsilon parameter ranging from 0.01 to 0.09, $k$ is the number of windows ranging from one to three, and $\beta_1$, $\beta_2$, and $\beta_3$ are associated parameters of the model.  $\beta_1 / 100$ shows the predicted change in annotations when $\epsilon$ increases by 0.01.  $\beta_2$ will show the predicted change in annotations when the number of windows increase by 1.

To conduct the experiment, the stopping point was found for each of the previously used categories of each dataset using different combinations of parameters with $\epsilon$ in the set \{0.01, 0.02, 0.03, 0.04, 0.05, 0.06, 0.07, 0.08, and 0.09\} and with $k$ in the set \{1, 2, 3\}.   For example, for the alt.atheism of 20Newsgroups, the number of annotations needed for stopping to occur were recorded for $\epsilon = 0.01$ and $k = 1$, $\epsilon = 0.02$ and $k = 1$, $\epsilon = 0.03$ and $k = 1$ and so on until each combination of parameter values were used.  Once the number of annotations for each category of each dataset was found, the data was trained on a linear regression model so that the independent variables were $\epsilon$ and $k$ and the dependent variable was the number of annotations.  The regressions were done for each dataset individually and then trained on all the datasets together.  The batch size was also kept constant throughout the experiments at 0.05\% of the training set.

The results of the experiment are shown in Table ~\ref{t:parameter_analysis_1}. Several conclusions can be drawn from Table~\ref{t:parameter_analysis_1}.  First, models that were made for specific datasets explained significantly more of the variation of annotations than when clustered all together.  This shows evidence that the change of annotations is dependent on the dataset. Second, for every increase in $\epsilon$ by 0.01, the number of annotations decreased on average anywhere from 259 to 2218 depending on the dataset used. This was also statistically significant and confirms our intuition.  When the window increased by one, the number of annotations on average needed increase from 531 to 1463.  This too was also statistically significant for every dataset.

\begin{table}[t]
\caption{This table shows the $R^2$, $\beta_1 / 100$, $\beta_2$, and $p$-values for $\beta_1$ and $\beta_2$ for each dataset done separately and all the datasets together.  $R^2$ explains what percentage of the variance of annotations is explained for by the variation of the independent variables of $\epsilon$ and $k$.  $\beta_1 / 100$ shows the change in annotations for every increase in $\epsilon$ by 0.01 from 0.01 to 0.09.  $\beta_2$ shows the change in annotations for every increase in the number of windows.  Letting $\alpha = 0.05$, if the $p$-values for each $\beta$ parameter is below $\alpha$, then the $\beta$ parameter is statistically significant.}
\label{t:parameter_analysis_1}
\centering
\begin{tabular}{ | c | c | c | c | c | c | }
\hline
Dataset & $R^2$ & $\beta_1 / 100$ & $\beta_2$ & $p_1$-val & $p_2$-val \\
\hline
20News & 44.7\% & -421.36 & 680.81 & 6.90e-62 & 3.32e-12 \\
\hline
Reuters & 42.2\% & -258.87 & 531.64 & 2.70e-26 & 6.15e-9 \\
\hline
WebKB & 55.1\% & -727.32 & 645.61 & 6.69e-192 & 2.37e-10 \\
\hline
ALL & 44.5\% & -572.98 & 639.39 & 1.46e-238 & 9.84e-21 \\
\hline
\end{tabular}
\end{table}

Another goal of this section was to see what type of relationship existed between the parameters and the difference between the actual change of F and the predicted change of F.  In Section ~\ref{h:AccuracyPredictingChangeF}, it was said that as the predicted change of F decreases, the prediction becomes more accurate.  In other words, as the method becomes more conservative, the predicted change of F measure becomes more accurate.  This observation will be tested by fitting a model of the form: $\hat{y} = \beta_1\epsilon + \beta_2k + \beta_3$ where $\hat{y}$ is the predicted difference between the actual change of F and the predicted change of F, $\epsilon$ is the epsilon parameter ranging from 0.01 to 0.09, $k$ is the number of windows ranging from one to three, and $\beta_1$, $\beta_2$, and $\beta_3$ are associated parameters of the model.  $\beta_1 / 100$ shows the predicted change in annotations when $\epsilon$ increases by 0.01.  $\beta_2$ will show the predicted change in annotations when the number of windows increase by 1.

The data was compiled the same way as the previous regression, except the dependent variable is now the difference between the actual change of F and the predicted change of F rather than the number of annotations.

The results of the experiment are shown in Table ~\ref{t:parameter_analysis_2}. There are several conclusions that can be drawn from the following table.  First, there is a statistically significant relationship between $\epsilon$ and $k$ and the difference between the predicted and actual change of F measure.  Second, the variance in $\epsilon$ and $k$ explains more of the variance of the difference between the actual and predicted change of F measure then the number of annotations.  Third, it can be said then that making $\epsilon$ smaller or increasing the number of windows can lead to a more accurate prediction.

\begin{table}[t]
\caption{This table shows the $R^2$, $\beta_1 / 100$, $\beta_2$, $\beta_3$, and $p$-values for $\beta_1$ and $\beta_2$ for each dataset done separately and all the datasets together.  $R^2$ explains what percentage of the variance of annotations is explained for by the variation of the independent variables of $\epsilon$ and the number of windows.  $\beta_1 / 100$ shows the change in the difference of the predicted and the actual change of F measure for every increase in $\epsilon$ by 0.01 from 0.01 to 0.09.  $\beta_2$ shows the change in the difference of the predicted and the actual change of F measure for every increase in the number of windows.  Letting $\alpha = 0.05$, if the $p$-values for each $\beta$ parameter is below $\alpha$, then the $\beta$ parameter is statistically significant.}
\label{t:parameter_analysis_2}
\centering
\begin{tabular}{ | c | c | c | c | c | c | c | }
\hline
Dataset & $R^2$ & $\beta_1 / 100$ & $\beta_2$ & $p_1$-val & $p_2$-val \\
\hline
20News & 51.4\% & 3.56e-3 & -6.13e-3 & 1.21e-71 & 2.86e-15 \\
\hline
Reuters & 43.7\% & 3.86e-3 & -5.35e-3 & 0.00 & 5.42e-12 \\
\hline
WebKB & 82.5\% & 6.99e-3 & -5.46e-3 & 0.00 & 6.03e-12 \\
\hline
ALL & 56.0\% & 5.57e-3 & -5.63e-3 & 0.00 & 9.25e-22 \\
\hline
\end{tabular}
\end{table}

\section{Conclusion}

A new stopping method was introduced called predicted change of F.  This stopping method provides the user the flexibility to adjust the parameters $\epsilon$ and $k$ so that the stopping method can be more or less aggressive.  This stopping method also provides an intuition of how much the performance of the model is changing.  Finally, this stopping method can be used with any base learner.

The intuition behind such a method is that we want to use a stop set that does not need to be labeled to predict what the change of performance is expected to be.  In order to predict the change of performance, we rely on an assumption that $M_t$ will be superior to $M_{t-1}$ because the most recent iteration would have been trained on a larger training set than the previous iteration.  This assumption was found in Section~\ref{h:Assumption} to be reasonable.  This assumption allows the predicted change of F to be evaluated as $\Delta \hat{F} = \frac{2(a + c)}{2(a + c)} - \frac{2a}{2a + c + b} = 1 - \frac{2a}{2a + c + b}$. As found in Section~\ref{h:Accuracy}, the predicted change of F was found to be closer to the true change of F on the stop set than the upper bound proven by Bloodgood and Grothendieck \cite{bloodgood2013}.  

We investigated the relationship between the parameters $\epsilon$ and $k$ and the number of annotations along with the accuracy of the predicted change of F measure.  Ultimately, both parameters were found to be statistically significant.  Furthermore, as expected, when $\epsilon$ decreased and the number of windows $k$ increased, the stopping method became more conservative and more annotations were required.  Furthermore, as $\epsilon$ decreased and $k$ increased, the prediction of the change of F measure became more accurate.

The predicted change of F was found to be very accurate as the number of annotations increased.  However, at earlier iterations, the predicted change of F measure was unreliable.  This is not a problem when using predicted change of F as a stopping method, as the method only stops when the predicted change of F converges to a small number.  The new stopping method was found to be effective due to the fact that it provides flexibility through its parameters, provides estimates of how much performance is changing, and can be applied with any base learner without any significant computational cost.

\section*{Acknowledgment}

This work was supported in part by The College of New Jersey Support of Scholarly Activities (SOSA) program and by The College of New Jersey Mentored Undergraduate Summer Experience (MUSE) program.  The authors acknowledge use of the ELSA high performance computing cluster at The College of New Jersey for conducting the research reported in this paper.  The cluster is funded by the National Science Foundation under grant number OAC-1828163.

\bibliographystyle{IEEEtran}
\bibliography{paper}

\end{document}